\documentclass{article}

\usepackage[preprint]{corl_2026}

\usepackage{amsmath}
\usepackage{amssymb}
\usepackage{graphicx}
\usepackage{booktabs}
\usepackage{algorithm}
\usepackage{algpseudocode}
\usepackage{wrapfig}
\usepackage{multirow}
\usepackage[table]{xcolor}
\usepackage{caption}
\usepackage{subcaption}
\usepackage{graphicx}

\title{Generating Robot Hands from Human Demonstrations}

\author{%
Sha Yi$^{1}$ \quad
Nicklas Hansen$^{1}$ \quad
Xueqian Bai$^{1}$ \quad
Carmelo Sferrazza$^{2}$ \\
\textbf{Michael T. Tolley$^{1}$ \quad
Xiaolong Wang$^{1}$} \\[0.3em]
$^{1}$University of California San Diego \quad
$^{2}$Amazon Frontier AI \& Robotics
}

\begin{document}
\maketitle

\begin{center}
\vspace{-10pt}
\captionsetup{type=figure}
\includegraphics[width=0.98\linewidth]{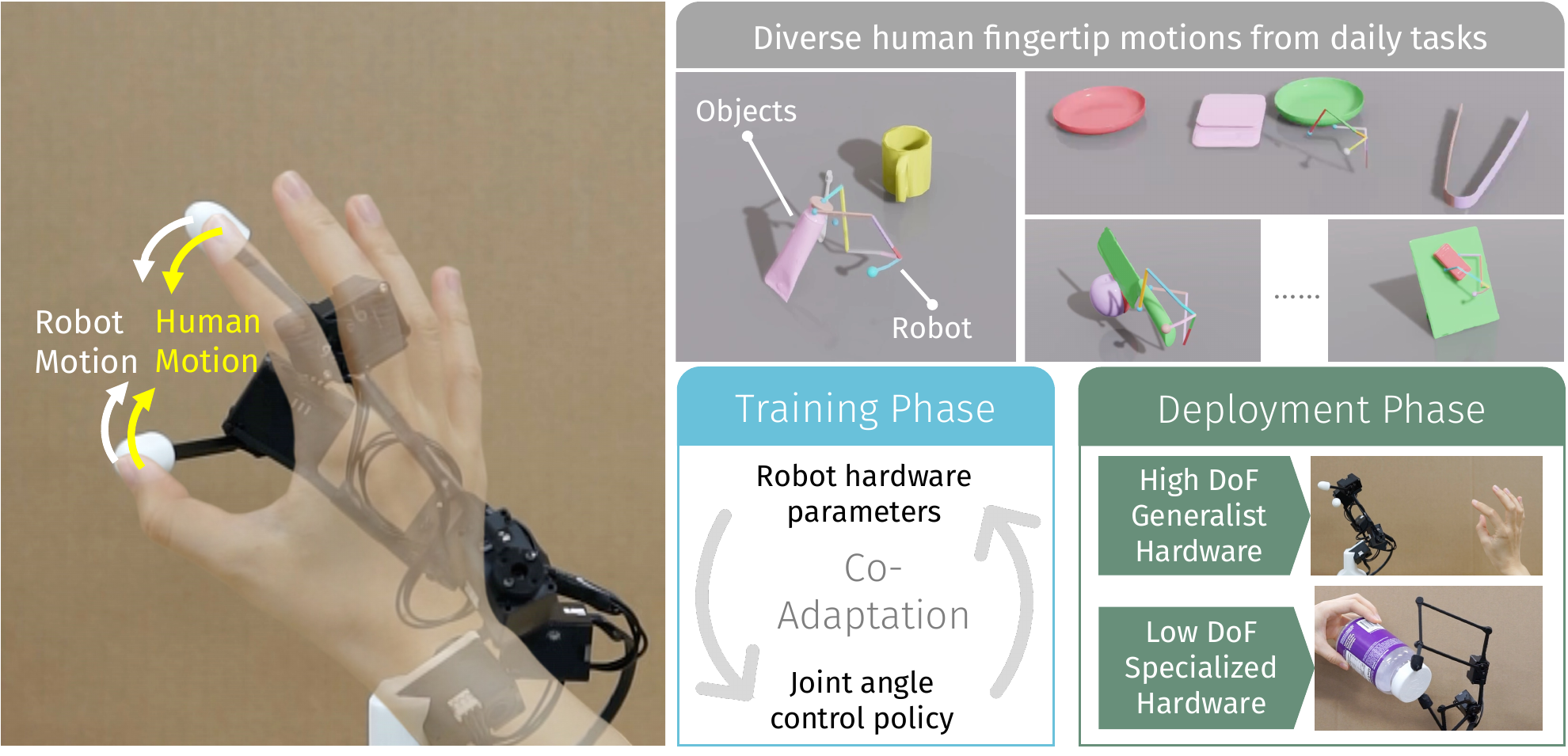}
\caption{\footnotesize We use diverse human hand motions from daily manipulation as targets for robot hand generation. During training, robot hardware parameters and the joint-angle control policy are optimized together to match the observed fingertip motions. We can produce either high-DoF generalist hardware for broad teleoperation or low-DoF specialized hardware for structured task trajectories.}
\label{fig:overview}
\end{center}

\begin{abstract}
Robot learning has advanced rapidly in learning control, but learning the physical body of a robot remains much more difficult because jointly searching over design and control creates a very large combinatorial problem. Here, we present a data-driven framework for generating robot hands from human demonstrations. Instead of learning a complex controller together with each candidate design, we generate robot hand designs using the same simple control policy used after fabrication: matching fingertip positions through inverse kinematics. Using more than 4 million frames of human fingertip motion from everyday manipulation, our algorithm optimizes tree-structured robot hands to reproduce desired target motions. The framework produced both a 6-degree-of-freedom (DoF) general-purpose hand and lower-DoF task-specific hands with spatial four-bar mimic joints. To accelerate the search over designs, we trained a reinforcement-learning (RL) actor to propose good hand designs and joint angles, reducing search time from hours to minutes. We fabricated the mechanisms directly as one-piece articulated structures with print-in-place joints. In real-world experiments, the 6-DoF hand achieved highly accurate teleoperated fingertip tracking better than available commercial robot hands, whereas the specialized 3-DoF hands reproduced structured human and synthetic trajectories with reduced mechanical complexity. These results showed that large-scale human motion data can be used not only to train robot controllers but also as a reference for optimizing and generating the physical embodiment of robots. Webpage: \url{https://yswhynot.github.io/generating-robot-hands/}
\end{abstract}


\section{Introduction}
\label{sec:introduction}

Robot learning has advanced rapidly in learning the \emph{brain} of robots. With large-scale data, teleoperation, and policy learning, robots can now acquire increasingly capable controllers for locomotion and manipulation~\citep{kumar2021rma,luo2025precise,zhao2023learning}. Yet in physical systems, intelligence does not reside in control alone. A robot acts through its body, and the body determines what motions are reachable, what contacts are stable, what forces can be transmitted, and which behaviors are easy or difficult to realize~\citep{pfeifer2006body}. If we want robots to perform useful physical work, we need methods not only for learning controllers, but also for generating embodiments.

This naturally suggests co-design: jointly optimizing hardware and control. Co-design is powerful, but it is also difficult because the design space and control space are coupled. Changing the geometry of a hand changes the controller that best fits a motion, and changing the controller changes which designs appear useful. This coupling creates a large, nonconvex search problem, especially when the goal is not a single scripted motion but a hand that can reproduce a broad class of manipulation behaviors.

We take advantage of an asymmetry between design and control. During training, both hardware parameters and joint trajectories can be optimized. At deployment, however, the hardware is fixed once fabricated, while the controller remains adjustable online. Therefore, if a simple controller will be used after fabrication, the design should be learned under that same controller. In this work, we optimize robot hands so that human thumb-index fingertip motions are reproducible under inverse kinematics, rather than learning a separate complex policy for every candidate design.

Human hand motion is a natural behavioral prior for this problem. Human demonstrations are diverse, abundant, and representative of the manipulation behaviors that robots are expected to perform. At the same time, human hands are mechanically difficult to replicate: practical robot hands must operate with far fewer actuators because of constraints on size, cost, wiring, robustness, and electronics integration. Retargeting can map human motions to an existing robot hand~\cite{qin2023anyteleop, yin2025geometric}, but it cannot remove the underlying kinematic mismatch introduced by the chosen embodiment. We instead use human fingertip trajectories to generate the embodiment itself.

We present a deployment-aligned co-design framework that optimizes tree-structured two-finger hands from more than 4 million frames of human manipulation data~\cite{zhan2024oakink2}. The framework optimizes both fully actuated hands and low-DoF hands with spatial four-bar mimic joints (Fig.~\ref{fig:overview}). This lets a single formulation produce a 6-DoF general-purpose hand that covers a broad distribution of human fingertip motions, as well as task-specific hands whose passive couplings encode structured motion directly into hardware. To make constrained low-DoF design generation efficient, we train a trajectory-conditioned actor that proposes hardware and joint-angle initializations, reducing hardware generation time from hours to minutes. In simulation and real-world experiments, the generated 6-DoF hand achieves sub-millimeter tracking error and accurate real-time teleoperation, while specialized 3-DoF hands track both human and synthetic trajectories with reduced mechanical complexity.

\section{Related Work}
\label{sec:related_work}

\textbf{Embodiment and robot co-design.}
Embodied agents show that behavior depends on both control and morphology~\citep{varela1991embodied,pfeifer2006body,pfeifer2007self,nolfi2002synthesis}. This view has motivated robot co-design methods that jointly optimize morphology, actuation, trajectories, and control~\citep{lipson2000automatic,young2003evolution,mintchev2016adaptive,spielberg2017functional,schulz2015interactive,ha2018computational,chen2021co,carter2023task,fletcher2025multidisciplinary,kodnongbua2023computational}, as well as methods that adapt morphology within policy learning~\citep{ha2019reinforcement,yuan2021transform2act,gupta2021embodied,islam2024task,he2024morph,li2024reinforcement}. These approaches are powerful, but often depend on task-specific rewards and repeated controller adaptation as the hardware changes~\citep{xu2021end,ha2021fit2form,xu2024dynamics,yi2025co,bai2025learning}. We instead optimize morphology under a fixed deployment-time control prior: inverse kinematics.

\textbf{Robot design spaces.}
Searchable design spaces have been explored through differentiable simulators, graph grammars, soft-body models, and structured mechanism generation~\citep{ma2021diffaqua,zhao2020robogrammar,liu2024paperbot,zhao2022automatic,obayashi2022soft}, enabling morphology search for locomotion, manipulation, modular robots, and other freeform robots~\citep{coros2013computational,luck2020data,schaff2022soft,polzin2025robotic,fay2025cross,pathak2019learning}. Our design space is narrower but fabrication-oriented: tree-structured two-finger linkages, fully actuated joints, and spatial four-bar mimic joints based on Bennett-linkage coupling~\citep{perez2002bennett}. For constrained low-DoF generation, we compare the actor-initialized refinement procedure against Cross-Entropy Method search~\citep{rubinstein2004cross}.

\textbf{Human motion for robot hands.}
Large-scale robot learning has made significant progress in locomotion and manipulation controllers~\citep{kumar2021rma,he2024omnih2o,cheng2024expressive,luo2025precise,chen2023visual,zhao2023learning,fu2024mobile}. Human hand motion is a key supervision source for teleoperation, retargeting, and imitation learning~\citep{sivakumar2022robotic,qin2023anyteleop,shaw2024learning,li2025maniptrans,qiu2025humanoid,kareer2025egomimic,jiang2025dexmimicgen}, but human hands are mechanically richer than practical robot hands, which must balance dexterity, size, cost, robustness, wiring, and integration~\citep{dollar2010highly,catalano2014adaptive,odhner2014compliant,khatib2016ocean,laffranchi2020hannes,bircher2021complex,shaw2023leap,zorin2025ruka,christoph2025orca,gilday2025embodied,capsi2025merging}. Retargeting maps human motion to a chosen hand~\citep{chong2021learning,yin2025geometric,lakshmipathy2025kinematic}. Thus, we use human motion data to learn the hand hardware itself.

\section{Method}
\label{sec:method}

\subsection{Problem Formulation}\label{sec:formulation}

As shown in Fig.~\ref{fig:overview}, the input to the framework is a collection of human thumb-index fingertip motions, and the output is either a general-purpose high-DoF hand or a low-DoF hand specialized to a target trajectory. We represent each target demonstration as a thumb-index fingertip trajectory in the wrist frame,
\begin{equation}
X^\star = \{x_t^\star\}_{t=1}^{T}, \qquad x_t^\star \in \mathbb{R}^{6},
\end{equation}
where \(x_t^\star\) contains the 3D positions of the two fingertips at time \(t\). A candidate robot hand is parameterized by hardware variables \(\phi\) and a joint-angle trajectory \(q\). We formulate the forward kinematics as a differentiable process that calculates fingertip positions based on the given parameters
\begin{equation}
\hat{X} = g(\phi, q).
\end{equation}
We solve for design and motion variables by minimizing
\begin{equation}
\Theta^\star =
\arg\min_{\phi,q}
\mathcal{L}_{\mathrm{track}}
+ \lambda_{\mathrm{joint}}\mathcal{L}_{\mathrm{joint}}
+ \lambda_{\mathrm{design}}\mathcal{L}_{\mathrm{design}}
+ \lambda_{\mathrm{col}}\mathcal{L}_{\mathrm{col}},
\label{eq:overall_loss}
\end{equation}
where \(\Theta=\{\phi,q\}\). The tracking error and the joint smoothness loss are
\begin{equation}
\mathcal{L}_{\mathrm{track}} =
\frac{1}{T}\sum_{t=1}^{T}\left\|\hat{x}_t - x_t^\star\right\|_1, \qquad
\mathcal{L}_{\mathrm{joint}} =
\frac{1}{T-1}\sum_{t=1}^{T-1}
\left\|q^{\mathrm{eff}}_{t+1}-q^{\mathrm{eff}}_{t}\right\|_2^2,
\end{equation}
The joint term penalizes rapid changes within each trajectory's sequence. The design term discourages unnecessarily long links and, for mimic-joint hands, regularizes the linkage parameters. The collision term uses centerline segments from forward kinematics based on the positions of link segments. For valid non-adjacent segment pairs \((i,j)\), it applies
\begin{equation}
\mathcal{L}_{\mathrm{col}} =
\sum_{(i,j)}
\max(0,w-d_{ij}),
\end{equation}
where \(d_{ij}\) is the closest distance between two link segments and \(w\) is a clearance radius. This gives a smooth approximation that is inexpensive enough to use inside repeated design optimization. Another loss not included in the equation is the joint limits based on the types of configuration (fully actuated or with mimic joints). Additional implementation details are given in Appendix~\ref{sec:appendix_optimization}.

\subsection{Design Space}

\begin{figure}[tb]
\centering
\includegraphics[width=0.9\linewidth]{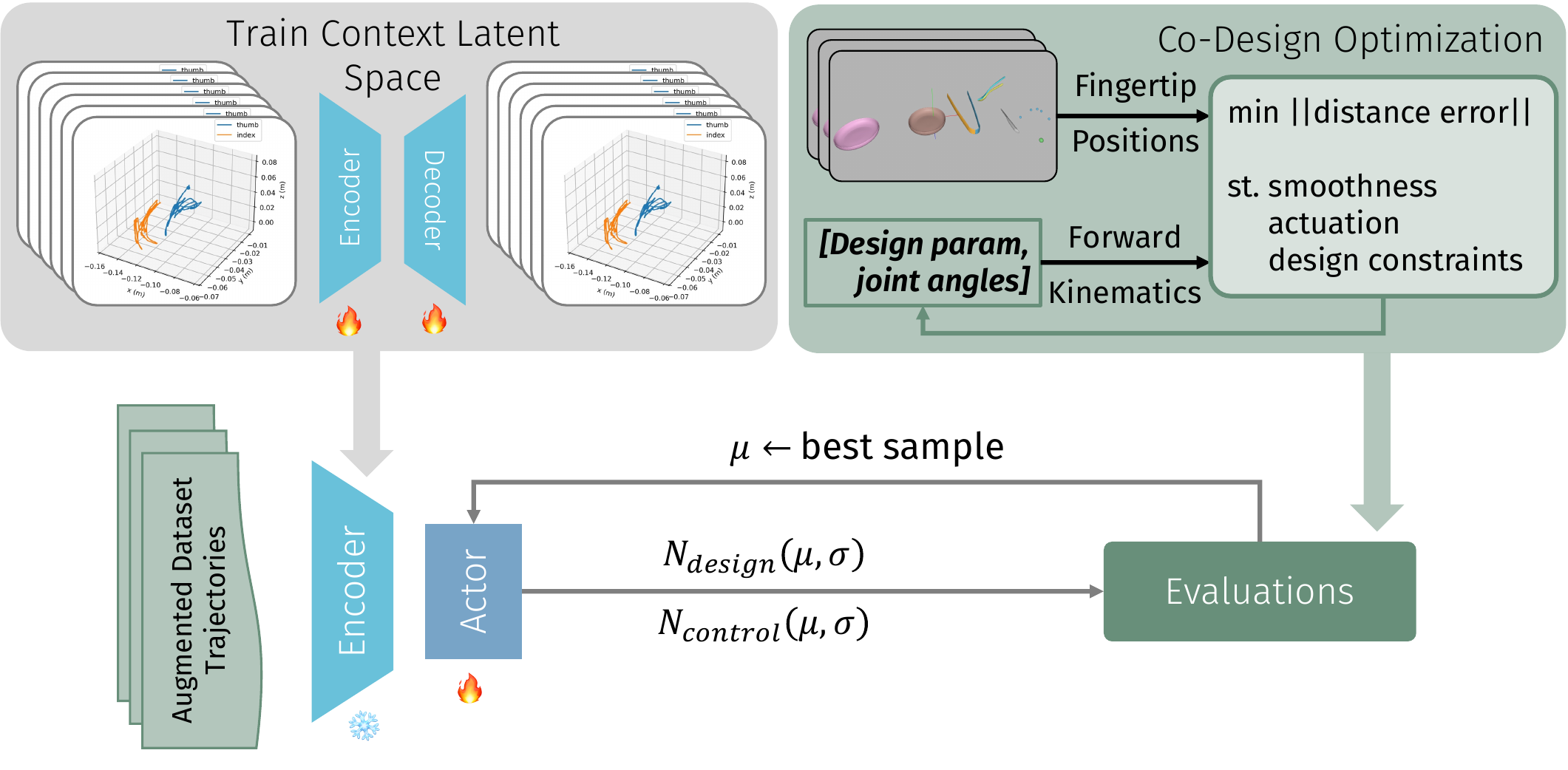}
\caption{\footnotesize\textbf{Training pipeline of trajectory-conditioned co-design.} A trajectory autoencoder is trained on augmented thumb-index trajectories to learn a compact motion context. The frozen encoder provides this context to an actor, which samples candidate design parameters and joint-angle initializations. Each candidate is evaluated by differentiable co-design optimization, which minimizes fingertip distance error subject to smoothness, actuation, and design constraints. The actor's mean is updated toward the best evaluated sample.}
\label{fig:method}
\vspace{-10pt}
\end{figure}
The hand is modeled as a tree-structured linkage rooted at the wrist, with two branches corresponding to the thumb and index fingertips. For fully actuated hands, \(\phi\) includes link lengths, pre-link lengths for motor mounting, and joint orientations. For lower-DoF hands, \(\phi\) also includes spatial four-bar mimic-joint parameters, namely Bennett joints~\cite{perez2002bennett}. Details are included in Appendix~\ref{sec:bennett}. These mimic joints passively couple child motion to parent motion, allowing a small number of actuators to generate structured trajectories that are out of the rotational plane.
We use a half-angle relation for a mimic joint as in~\cite{perez2002bennett},
\begin{equation}
\theta_c =
f - 2\,\operatorname{atan2}\!\left(
k\sin\frac{\theta_p}{2},
\cos\frac{\theta_p}{2}
\right),
\end{equation}
where \(\theta_p\) is the parent joint angle, \(\theta_c\) is the coupled child angle, and \(f,k\) are determined by optimizable linkage parameters related to the formulation of the Bennett spatial four-bar. The \(k\) determines how the child angle changes based on the input parent angle, and \(f\) is an angle offset. During gradient-based optimization, we use a softened residual parameterization rather than enforcing exact Bennett four-bar linkage closure. This keeps the induced motion close to the mimic joint constraints while avoiding the narrow feasible set of hard closed-chain constraints. After optimization, the remaining spatial four-bar geometry is recovered by nonlinear least squares for fabrication. Appendix~\ref{sec:bennett} gives the full derivation and synthesis procedure.

\subsection{Trajectory-Conditioned Hardware Generation}

For fully actuated hands, the formulation in Section~\ref{sec:formulation} remains end-to-end differentiable by representing joint rotations with the continuous 6D representation~\cite{zhou2018continuity}. This allows hardware parameters and joint angles to be co-optimized directly with gradient descent (GD) in a single unified framework.
The main difficulty arises when extending this differentiable optimization to hardware designs with Bennett link-closure constraints for mimic joints. These constraints introduce a highly nonconvex search space: passive couplings not only reduce the number of independent degrees of freedom, but also reshape the reachable set of the hand. As a result, naive GD becomes highly sensitive to initialization.

To address this issue, we amortize the initialization search with a trajectory-conditioned actor. Rather than starting GD from random hardware and joint angle configurations, the actor proposes candidate initializations conditioned on the target fingertip motion. A frozen trajectory encoder first maps the target motion to a compact context vector, and the actor uses this context to sample promising hardware parameters and joint-angle seeds. These samples are then refined by the differentiable GD optimizer. Figure~\ref{fig:method} summarizes this actor-based generation loop, which can be interpreted as a reinforcement-learning-style sampling method for learning effective initializations for subsequent gradient-based co-optimization. We first learn a trajectory encoder that maps the target motion to a compact context vector $z = E_\psi(X^\star)$.
The actor predicts the mean of a Gaussian over candidate hardware-control initializations,
\begin{equation}
\mu_\theta(z) = A_\theta(z), \qquad
a_k = \mu_\theta(z) + \sigma\epsilon_k,\quad \epsilon_k \sim \mathcal{N}(0,I).
\end{equation}
Each sampled action \(a_k\) is decoded into design parameters and joint-angle initialization, then refined for a limited number of differentiable co-design steps using Eq.~\ref{eq:overall_loss}. We score the refined candidates using a reward that combines final tracking loss, linkage collision penalty, and angle-consistency penalty as in Appendix Algorithm~\ref{alg:best_of_k_actor}:
\begin{equation}
r_k =
s^{\mathrm{col}}_k
s^{\mathrm{angle}}_k
s^{\mathrm{loss}}_k,
\end{equation}
where each score is a sigmoid-normalized preference for low violation or low tracking error. The best candidate becomes the supervised target:
\begin{equation}
k^\star = \arg\max_k r_k,\qquad
\mathcal{L}_{\mathrm{actor}} =
\left\|\mu_\theta(z)-a_{k^\star}\right\|_2^2.
\end{equation}
The actor is a 3-layer Multilayer perceptron (MLP). At test time, the actor outputs an initialization for a new target trajectory, followed by GD refinement with a small number of iterations. This turns hardware generation from a long trajectory-specific search into a reusable learned proposal-and-refinement loop.

\subsection{Fabrication and Deployment}

\begin{figure}[t]
\centering
\includegraphics[width=0.98\linewidth]{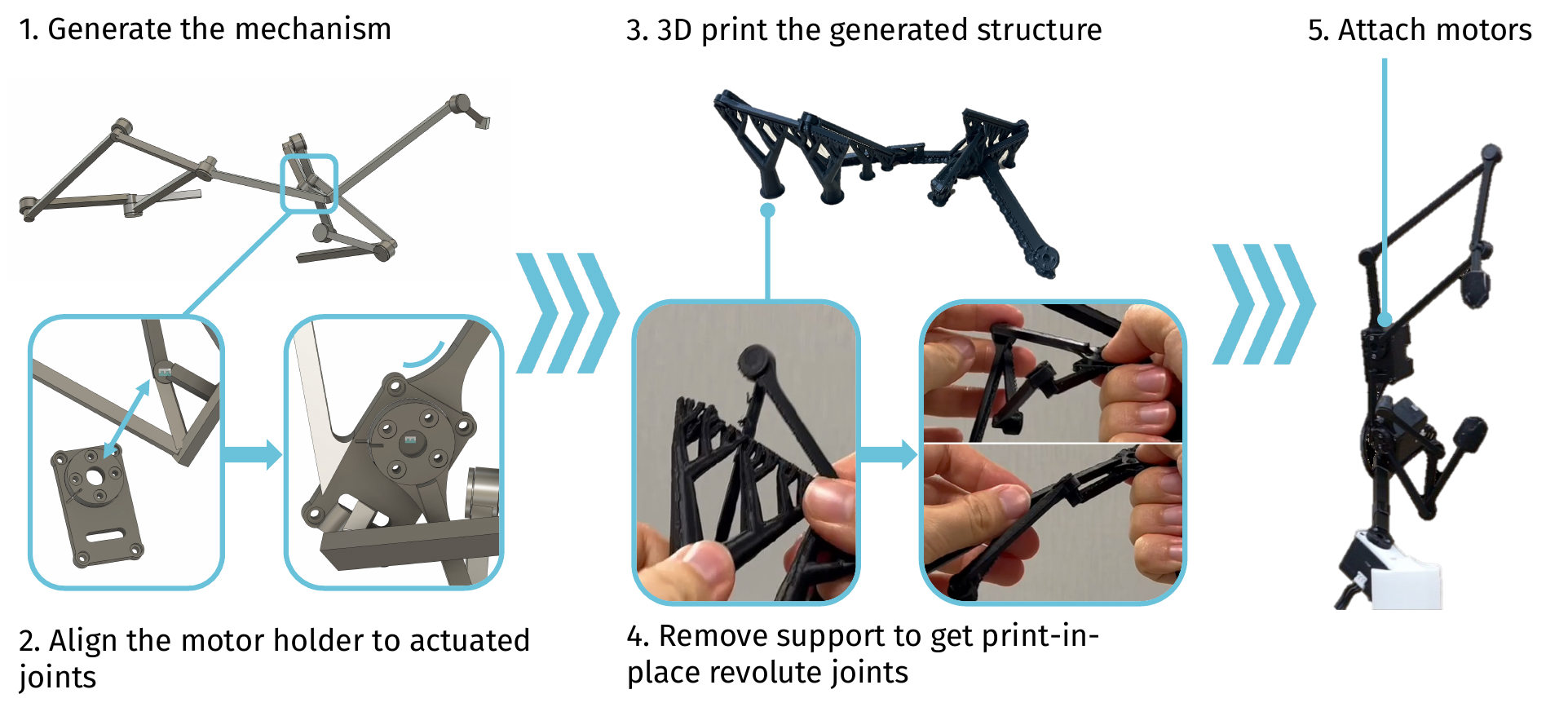}
\caption{\footnotesize\textbf{Fabrication workflow.} The optimized mechanism is generated directly as meshes consisting of boxes, rings, and cylinders. Motor holders are aligned to the actuated joints, and the structure is directly 3D printed as a single piece. After support removal, the print-in-place revolute joints articulate, and motors are attached to produce the final robot hand.}
\label{fig:manufacturing}
\vspace{-10pt}
\end{figure}

Optimized kinematic designs are converted into physical mechanisms by generating rigid links, joint elements, and mimic-linkage components as meshes (Fig.~\ref{fig:manufacturing}). The links are boxes, and each joint consists of a cylinder, two discs at each end, and a ring centered at the joint. Motor holders are aligned to actuated joints, and the design is 3D-printed as a single print-in-place structure on a tabletop 3D printer, as in recent progress about accessible 3D mechanism fabrication~\citep{zhai2023desktop,conrad20243d,liu2025advances}. After support removal, the revolute joints can rotate in place without separate assembly. 

\section{Experiments}
\label{sec:experiments}

\begin{figure}[t]
\centering
\includegraphics[width=0.98\linewidth]{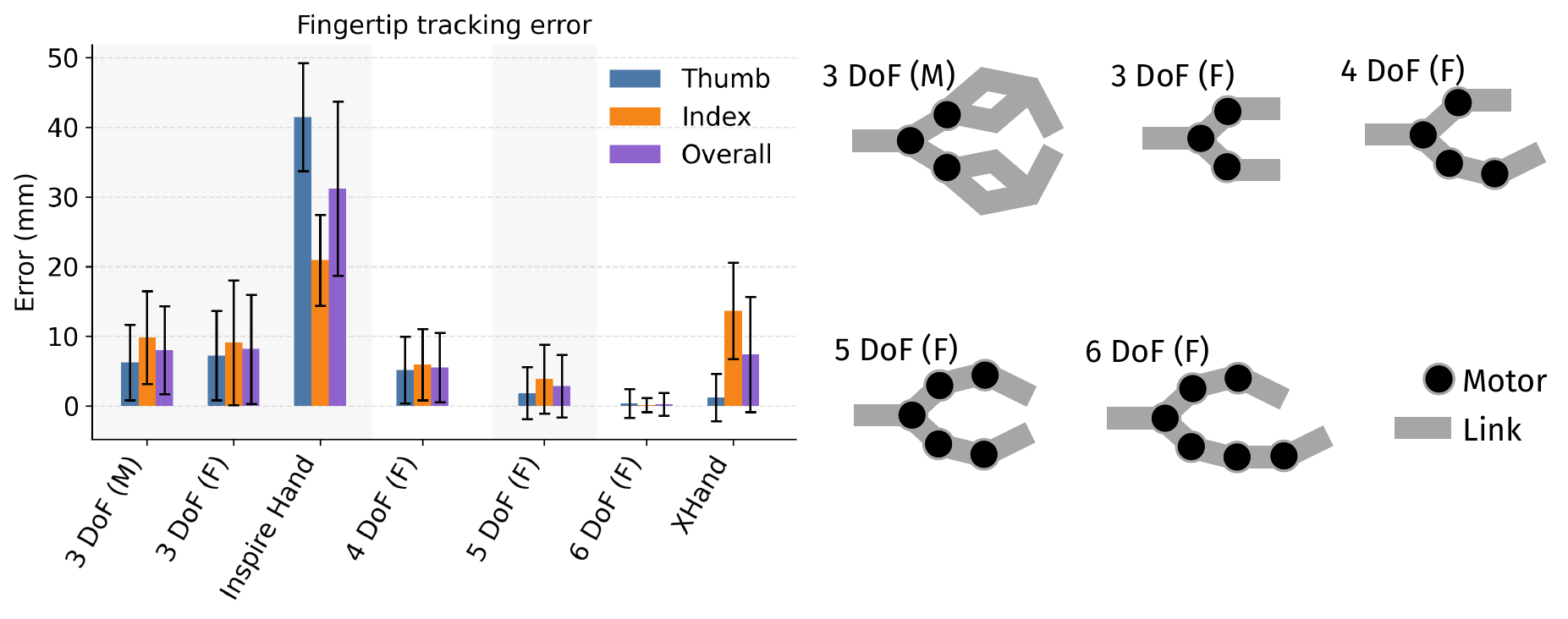}
\vspace{-10pt}
\caption{\footnotesize \textbf{Fingertip tracking error across hand designs.} Mean thumb, index, and overall fingertip tracking error for generated hands and commercial baselines on the full human motion dataset. Error bars denote one standard deviation across frames. The schematic on the right shows the generated kinematic structures and motor placements for different DoF settings.}
\label{fig:tracking_results}
\vspace{-10pt}
\end{figure}

\paragraph{Dataset and Evaluation}
We use OakInk human manipulation demonstrations~\citep{yang2022oakink,zhan2024oakink2}, containing 627 sequences and more than four million frames of daily tabletop and household manipulation tasks. We evaluate generated hands by the mean thumb-index fingertip tracking error and by the fraction of frames tracked within \(1\) mm. We compare generated hands with different DoF counts, a 3-DoF generated mimic-joint hand, and two commercial robot hand baselines: the Inspire Hand and XHand. Figure~\ref{fig:tracking_results} summarizes the mean tracking-error comparison, while Table~\ref{tab:tracking_tables} reports the tracking coverage and task-specific errors of low-DoF hands.

\begin{table}[tb]
\centering
\caption{\footnotesize \textbf{Tracking coverage and task-specific low-DoF errors.} Left: percentage of frames within \(1\) mm fingertip error on the full human motion dataset. Right: mean \(\pm\) standard deviation fingertip tracking error in millimeters for representative task-specific trajectories.}
\label{tab:tracking_tables}
\vspace{5pt}
\begin{minipage}[t]{0.36\linewidth}
\centering
\textbf{Full-dataset coverage}

\scriptsize
\setlength{\tabcolsep}{3.2pt}
\renewcommand{\arraystretch}{1.12}
\resizebox{\linewidth}{!}{%
\begin{tabular}{clcc}
\toprule
\textbf{DoF} & \textbf{Hand} & \textbf{Thumb} & \textbf{Index} \\
\midrule

& 3-DoF mimic & 12.22\% & 2.39\% \\
\textbf{3} & 3-DoF full & 10.07\% & 4.98\% \\
& Inspire Hand & 0.00\% & 0.04\% \\

\addlinespace[2pt]
\rowcolor{gray!10}
\textbf{4} & 4-DoF full & 16.27\% & 10.60\% \\

\addlinespace[2pt]
\textbf{5} & 5-DoF full & 63.12\% & 40.56\% \\

\addlinespace[2pt]
\rowcolor{gray!10}
& XHand & 83.69\% & 3.77\% \\
\rowcolor{gray!10}
\multirow{-2}{*}{\cellcolor{gray!10}\textbf{6}} 
& 6-DoF full & \textbf{95.38}\% & \textbf{98.19}\% \\
\bottomrule
\end{tabular}
}

\end{minipage}
\hfill
\begin{minipage}[t]{0.625\linewidth}
\centering
\textbf{Task-specific error}

\scriptsize
\setlength{\tabcolsep}{2.2pt}
\renewcommand{\arraystretch}{1.32}
\resizebox{\linewidth}{!}{%
\begin{tabular}{llccc}
\toprule
\textbf{Task} & \textbf{Hand} & \textbf{Thumb} & \textbf{Index} & \textbf{Overall} \\
\midrule

\rowcolor{gray!10}
& Mimic & 1.888 $\pm$ 2.257 & 2.784 $\pm$ 2.775 & 2.336 $\pm$ 2.569 \\
\rowcolor{gray!10}
\multirow{-2}{*}{\cellcolor{gray!10}Lid-off}
& Full & \textbf{1.457} $\pm$ 1.903 & \textbf{2.535} $\pm$ 2.688 & \textbf{1.996} $\pm$ 2.390 \\

\addlinespace[1pt]
& Mimic & \textbf{2.031} $\pm$ 1.694 & \textbf{0.174} $\pm$ 0.256 & \textbf{1.102} $\pm$ 1.526 \\
\multirow{-2}{*}{Key}
& Full & 2.282 $\pm$ 1.756 & 3.583 $\pm$ 2.690 & 2.933 $\pm$ 2.362 \\

\addlinespace[1pt]
\rowcolor{gray!10}
& Mimic & 0.015 $\pm$ 0.005 & \textbf{1.295} $\pm$ 0.960 & \textbf{0.655} $\pm$ 0.933 \\
\rowcolor{gray!10}
\multirow{-2}{*}{\cellcolor{gray!10}Circle-square}
& Full & \textbf{0.009} $\pm$ 0.002 & 10.851 $\pm$ 4.477 & 5.430 $\pm$ 6.278 \\

\bottomrule
\end{tabular}
}
\end{minipage}
\vspace{-10pt}
\end{table}

\paragraph{General-Purpose Hand Generation}

\begin{figure}[tb]
\centering
\includegraphics[width=0.98\linewidth]{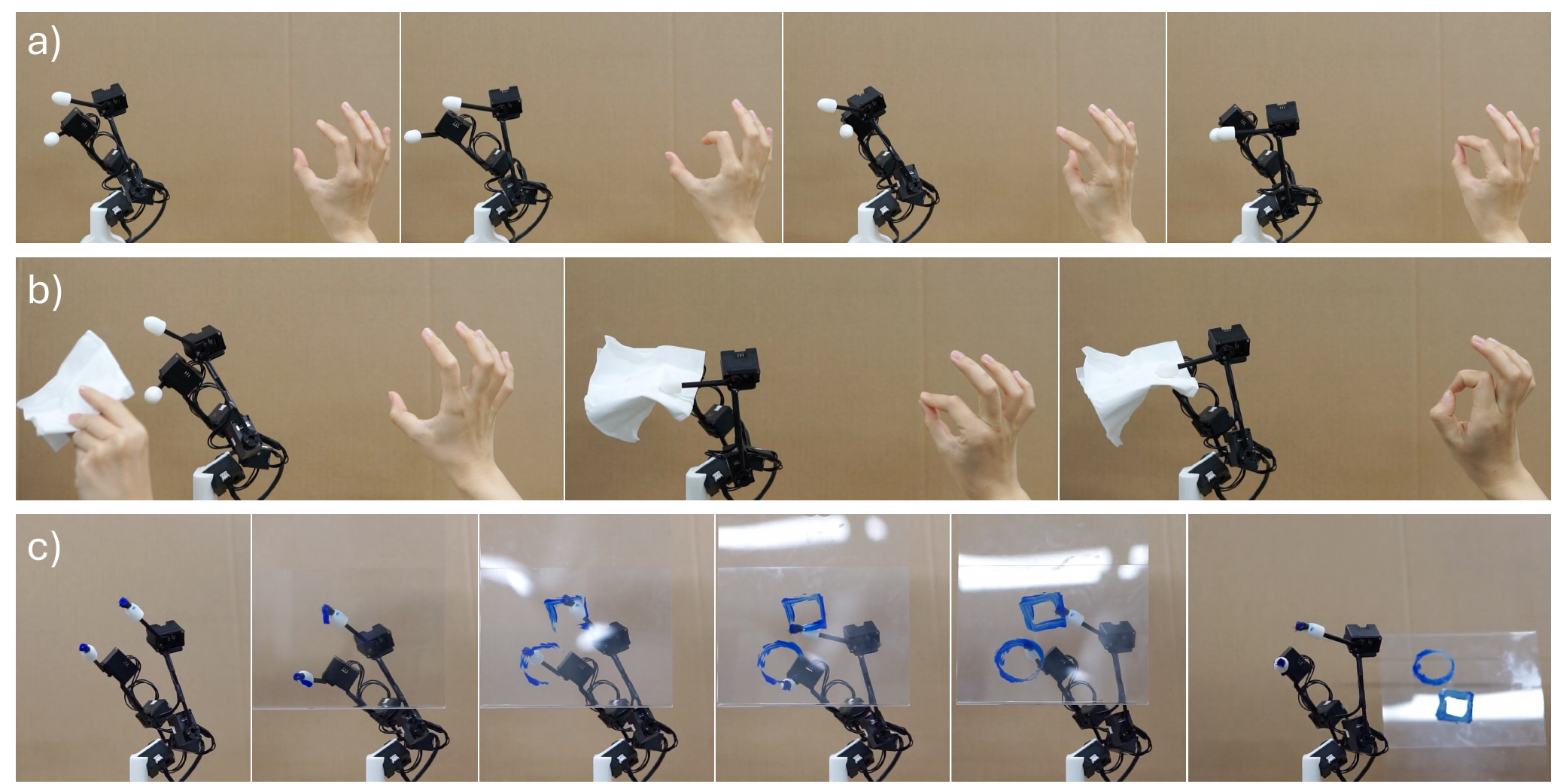}
\vspace{-5pt}
\caption{\footnotesize \textbf{Generated 6-DoF general-purpose hand.} (a) Real-time teleoperation using human thumb-index fingertip motion. The generated hand tracks a range of open, flexed, and pinch-like gestures. (b) Teleoperated pinch grasping and lifting of a thin napkin. (c) Programmed fingertip drawing in which the thumb traces a circle while the index finger traces a square.}
\label{fig:fulldof}
\vspace{-10pt}
\end{figure}

We first optimize a fully actuated hand over the complete human motion dataset. The resulting 6-DoF mechanism has one root joint, two joints on one branch, and three joints on the other branch. Although the optimizer solves for both design and joint angles during training, the learned joint angles are not directly used as deployment policies. The output is a hand geometry whose kinematics admit accurate inverse-kinematics solutions across the target motion distribution.

The 6-DoF hand achieves \(0.24\) mm overall mean fingertip error on the dataset, with \(0.11\) mm error for the index fingertip (Fig.~\ref{fig:tracking_results}). The improvement with DoF is highly nonlinear: generated 3-, 4-, and 5-DoF fully actuated hands obtain \(8.14\), \(5.53\), and \(2.84\) mm overall error, respectively. The final degree of freedom resolves a kinematic bottleneck in jointly positioning the two fingertips. The generated 6-DoF hand also reaches \(95.38\%\) of thumb frames and \(98.19\%\) of index frames within \(1\) mm, while the 5-DoF hand drops to \(63.12\%\) and \(40.56\%\) (Table~\ref{tab:tracking_tables}, left).
Commercial robot hands show that DoF count alone is not sufficient. The XHand also has 6 DoFs but obtains \(7.40\) mm overall error, with \(13.61\) mm index error. The Inspire Hand obtains \(31.17\) mm overall error. These comparisons indicate that the advantage comes from shaping the hand hardware design to fit the target motion distribution, rather than simply increasing the number of DoFs.

Real-world demonstrations support the same conclusion (Fig.~\ref{fig:fulldof}). Under real-time teleoperation, the generated hand tracks open-hand, partially flexed, and pinch-like gestures. It can pinch and move a thin napkin when the teleoperator maintains fingertip contact. The robot hand is also capable of producing motions beyond teleoperation, for example a sequence of programmed motions such as drawing a circle with one fingertip while drawing a square with the other. Such a task is very challenging for a human teleoperator, but since it lies within the workspace of a hand, a robot hand can be easily programmed to do so.

\paragraph{Low-DoF Task-Specialized Hands}

\begin{figure}[tb]
\centering
\includegraphics[width=0.98\linewidth]{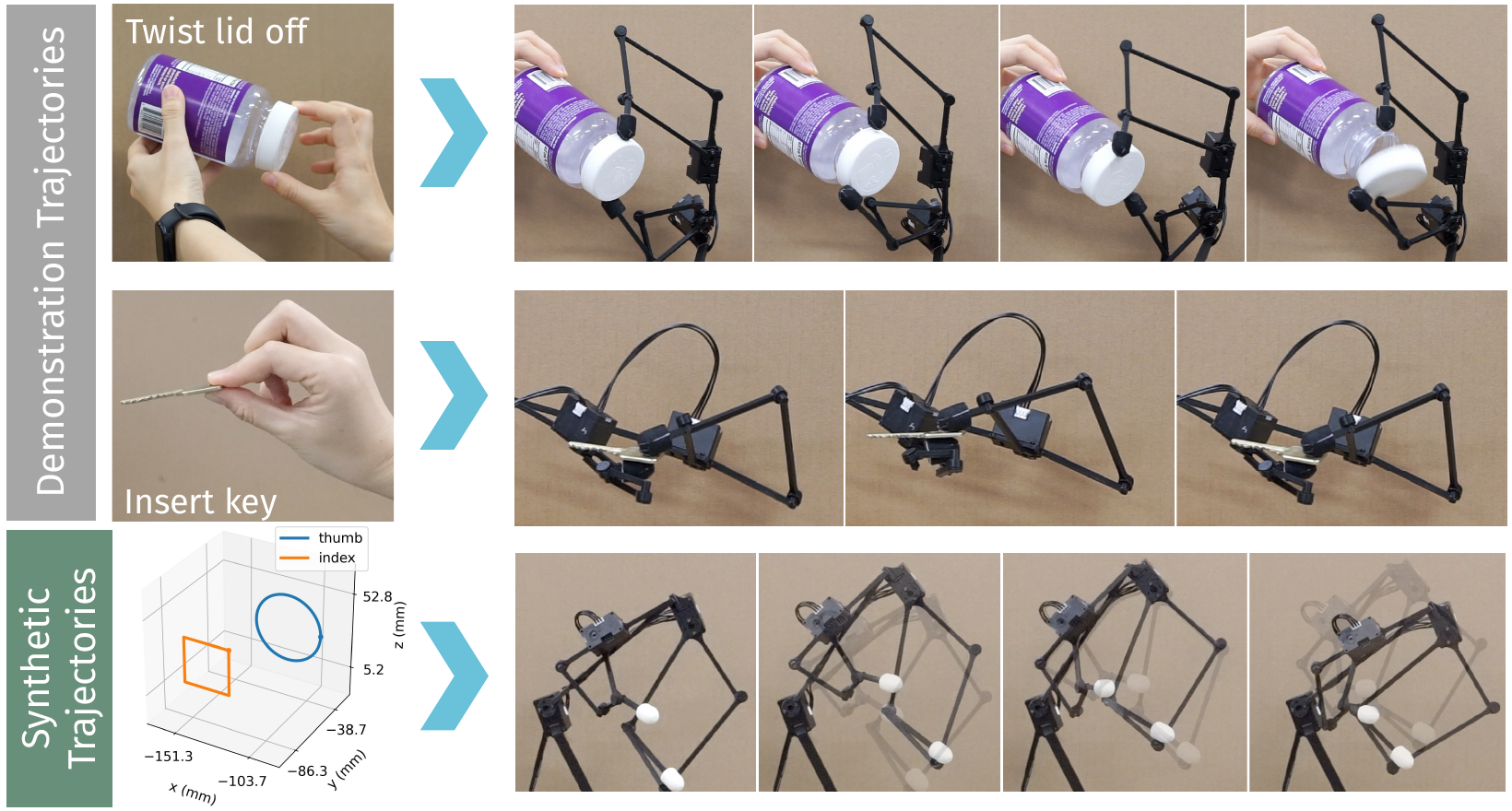}
\vspace{-5pt}
\caption{\footnotesize \textbf{Generated 3-DoF task-specialized hands.} Human demonstration and synthetic trajectories are used to generate low-DoF hands with spatial four-bar mimic joints. Top: a lid-twisting motion produces a hand that rotates the lid. Middle: a key-insertion motion produces a hand that holds and inserts the key. Bottom: a synthetic circle–square trajectory yields a specialized mechanism that reproduces the structured motion.}
\label{fig:mimic}
\vspace{-10pt}
\end{figure}

We next evaluate 3-DoF task-specialized hands with spatial four-bar mimic joints. These hands trade broad dexterity for reduced actuation, wiring, weight, and cost. We generate hands for lid twisting, key insertion, and a synthetic circle-square fingertip trajectory (Fig.~\ref{fig:mimic}). Across all three, the synthesized mechanisms reproduce the target behavior with only three actuated degrees of freedom.

The value of mimic joints depends on the trajectory geometry (Table~\ref{tab:tracking_tables}, right). On lid-off motion, the fully actuated and mimic 3-DoF hands perform similarly, with \(2.00\) mm and \(2.34\) mm overall error, because the target motion is close to planar circular motion. On key insertion, the mimic design improves error from \(2.93\) mm to \(1.10\) mm. On the circle-square trajectory, the mimic design reduces error from \(5.43\) mm to \(0.66\) mm by encoding non-circular index motion through the passive linkage. Thus, under a fixed actuator budget, structured passive kinematics can outperform a purely serial chain when the target motion has matching geometric regularity.

\paragraph{Actor-Based Search Acceleration}

\begin{figure}[tb]
\centering
\includegraphics[width=0.95\textwidth]{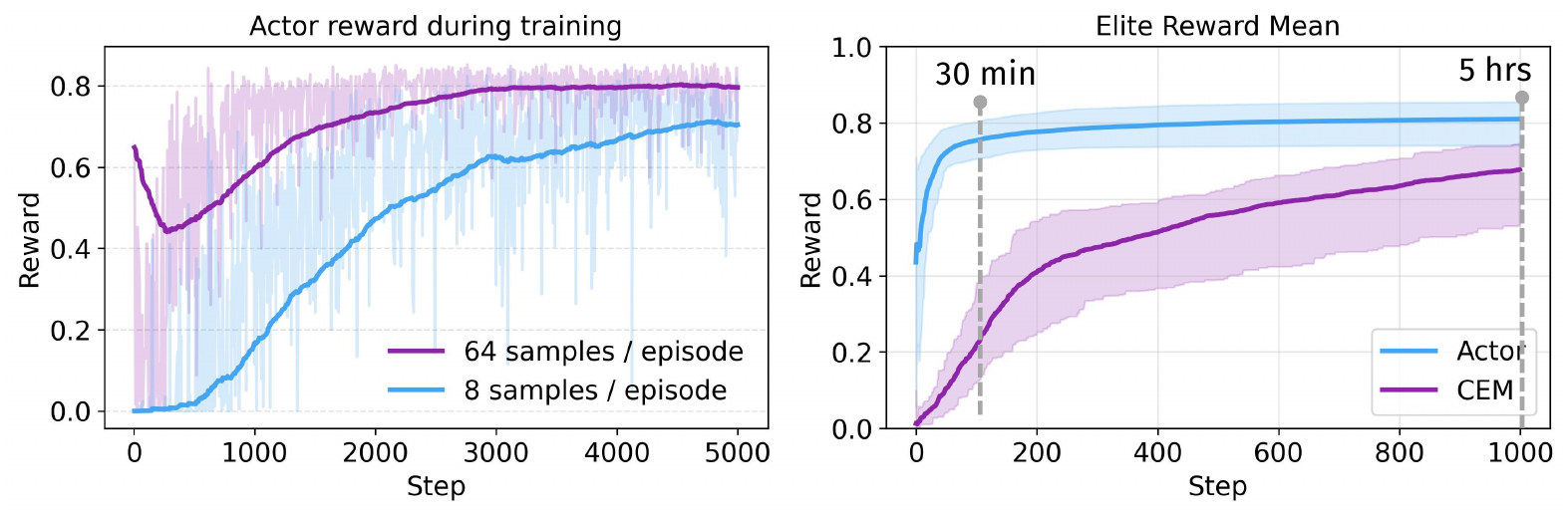}
\vspace{-5pt}
\caption{\footnotesize \textbf{Actor training and test-time generation efficiency.} Left: actor reward during training with 8 and 64 sampled candidates per episode. Right: Test-time elite reward comparing actor-initialized generation with trajectory-specific CEM. Actor initialization reaches a high-quality design within 30 minutes, whereas pure CEM fails to reach comparable performance after 5 hours.}
\label{fig:actor_efficiency}
\vspace{-10pt}
\end{figure}

Finally, we evaluate the trajectory-conditioned actor for hardware generation. During actor training, increasing the number of samples per episode from 8 to 64 improves the reward trajectory, but takes a significantly longer time (Fig.~\ref{fig:actor_efficiency}, left). At test time, standard Cross-Entropy Method (CEM) search~\citep{rubinstein2004cross} is expensive in the low-DoF mimic-joint space because each candidate requires constrained mechanism evaluation. The learned actor reaches a high elite reward in approximately 30 minutes, outperforming the design obtained by about 5 hours of trajectory-specific CEM (Fig.~\ref{fig:actor_efficiency}, right). This order-of-magnitude reduction makes task-specific embodiment generation practical for iterative design rather than a one-off offline procedure.

\section{Conclusion and Limitations}
\label{sec:conclusion_limitation}
We presented a framework for generating robot hands from human demonstrations. The central idea is to use human fingertip motion not only as a target for control, but also as a signal for shaping the hardware design itself. By optimizing hand kinematics under the same inverse-kinematics controller used after fabrication, the method searches for embodiments whose structure makes the desired motions easy to realize at deployment. This produces a 6-DoF general-purpose hand with broad thumb-index motion coverage, as well as low-DoF task-specialized hands whose spatial four-bar mimic joints encode structured motion through passive coupling. The actor-based initialization further shows that hardware search can be amortized across demonstrations, making task-specific mechanism generation substantially faster than solving each design problem from scratch.

\textbf{Limitations.} First, it optimizes only thumb-index fingertip positions. While this is useful for studying kinematic coverage and teleoperation, many manipulation tasks depend on full-hand interaction: contact over the fingers and palm, object geometry, compliance, friction, and the direction and magnitude of forces applied to the object. Second, the design space is restricted to two-finger tree-structured mechanisms and spatial four-bar mimic joints. This keeps the search tractable, but excludes richer palm geometries, contact surfaces and geometries, and other mechanism families. Third, the fabrication pipeline is not yet fully automatic. The generated meshes still require some manual processing, such as removing fused joints, checking clearances, and attaching or adjusting motor holders. These issues are mostly engineering limitations, but they currently affect the reliability and speed of moving from an optimized design to a working hand hardware prototype. Finally, the printed mechanisms are not yet strong enough for heavy manipulation tasks of high load, because the print-in-place joints can wear or break.

Future work would extend the objective beyond fingertip position tracking to include contact forces, object geometries and interactions, and also load. It should also expand the mechanism library, improve automatic CAD and mesh processing, and incorporate structural strength and manufacturability more into the optimization. All these efforts would bring such generated hardware into a broader range of real-world manipulation tasks.

\clearpage
% \acknowledgments{
% We thank Jianglong Ye and Yichen Zhai for helpful discussions. This work was supported, in part, by NSF CAREER Award IIS-2240014, NSF CCF-2112665 (TILOS), and gifts from Amazon and Meta.
% }

\bibliography{references}

\clearpage
\appendix
\section{Appendix}
\label{sec:appendix}
\subsection{Spatial four-bar linkage formulation}
\label{sec:bennett}

A classical Bennett linkage~\cite{perez2002bennett} is a spatial 4R overconstrained mechanism that consists of four revolute joints with a set of geometric constraints. Commonly, for robot hands, a four-bar linkage that has an actuated joint and a passively actuated joint is called \textit{mimic joint}. If \(d_i\) denotes the shortest distance between the two joint axes on link \(i\), and \(\alpha_i\) denotes the corresponding twist angle, then a movable Bennett 4R linkage satisfies
\begin{equation}
d_1=d_3,\qquad \alpha_1=\alpha_3,\qquad
d_2=d_4,\qquad \alpha_2=\alpha_4,
\end{equation}
together with the Bennett ratio condition
\begin{equation}
\frac{d_1}{\sin \alpha_1}=\frac{d_2}{\sin \alpha_2}.
\end{equation}
Under these constraints, the input and output joint angles follow a half-angle relation of the form
\begin{equation}
\tan\frac{\theta_c}{2}=k \tan\frac{\theta_p}{2},
\end{equation}
where \(k\) is determined purely by the linkage geometry. We can add an offset \(f\) if needed, as in:
\begin{equation}
\theta_c
=
f-2\arctan\!\left(k\tan\frac{\theta_p}{2}\right).
\end{equation}
This gives the characteristic nonlinear coupling between a parent joint angle \(\theta_p\) and its Bennett-coupled child angle \(\theta_c\).

For numerical optimization, we use the equivalent \texttt{atan2} form,
\begin{equation}
\theta_c
=
f-2\,\operatorname{atan2}\!\left(
k\sin\frac{\theta_p}{2},
\cos\frac{\theta_p}{2}
\right),
\end{equation}
which is algebraically equivalent to the half-angle form above but is more stable near angle wraparound.

Gradient-based design search benefits from smooth constraints. We therefore do \emph{not} enforce exact Bennett closure during gradient-based optimization. The hard Bennett equalities create a narrow, highly coupled feasible set that is sensitive to axis geometry. Instead, we use a Bennett-inspired soft constraint in which the child angle is still generated by a half-angle mimic relation, but the geometric coupling is relaxed with an optimizable residual.

Specifically, for mimic joint \(j\), we define
\begin{equation}
\theta^{c}_{j}
=
f_j
-2\,\operatorname{atan2}\!\left(
k_j \sin\frac{\theta^{p}_{j}}{2},
\cos\frac{\theta^{p}_{j}}{2}
\right),
\end{equation}
with
\begin{equation}
k_j=\frac{1}{\sin\tau_j + r_j}.
\end{equation}
Here, the optimizable parameters include: \(\tau_j\) is the parent-child axis skew term formulated as the joint angle, \(f_j\) is an angle offset, and \(r_j\) is a residual that relaxes the exact Bennett relation. Rather than forcing the true Bennett compatibility exactly, we only require
\begin{equation}
r_j \ge 0,
\end{equation}
implemented in practice by clamping the residual to a small positive lower bound during gradient descent. This keeps the denominator well-defined and preserves a Bennett-like soft constraint, while avoiding the ill-shaped hard-constraint landscape of the exact formulation.

We additionally use soft penalties on the induced effective mimic geometry, rather than exact equality constraints. This lets the optimizer exploit Bennett-like coupling when useful, while still exploring nearby designs that would be infeasible under the exact closed-form Bennett constraints. In other words, the formulation keeps the functional benefit of a Bennett mimic joint, but replaces the hard mechanism synthesis constraints with a smoother residual parameterization better suited to first-order optimization.
\subsection{Optimization framework}
\label{sec:appendix_optimization}

\begin{algorithm}[t]
\caption{\textsc{CoDesignOptimize}: Trajectory fitting with joint + design optimization}
\label{alg:joint_design_opt}
\begin{algorithmic}[1]
\Require target contact trajectory $\mathbf{X}^{\star}=\{\mathbf{x}^{\star}_t\}_{t=1}^T$, initial design parameters $\phi$, initial joint angles $\mathbf{q}$, learning rates $\eta_q,\eta_\phi$, number of iterations $K$
\Ensure optimized joint trajectory $\mathbf{q}$ and design parameters $\phi$

\For{$k = 1$ to $K$}
    \State Zero all gradients
    \State $\hat{\mathbf{X}}, \mathcal{S}, \mathbf{q}_{\mathrm{eff}} \gets \textsc{ForwardKinematics}(\mathbf{q}, \phi)$
    \Comment{$\hat{\mathbf{X}}$: predicted fingertip trajectory; $\mathcal{S}$: link segments}

    \State $\mathcal{L}_{\mathrm{track}} \gets \mathrm{Distance}\!\left(\hat{\mathbf{X}}, \mathbf{X}^{\star}\right)$
    \State $\mathcal{L}_{\mathrm{joint}} \gets \textsc{JointRegularization}(\mathbf{q}_{\mathrm{eff}})$
    \State $\mathcal{L}_{\mathrm{design}} \gets \textsc{DesignRegularization}(\phi)$
    \State $\mathcal{L}_{\mathrm{col}} \gets \textsc{SegmentCollisionPenalty}(\mathcal{S})$

    \State $\mathcal{L} \gets
    \lambda_{\mathrm{track}} \mathcal{L}_{\mathrm{track}}
    + \lambda_{\mathrm{joint}} \mathcal{L}_{\mathrm{joint}}
    + \lambda_{\mathrm{design}} \mathcal{L}_{\mathrm{design}}
    + \lambda_{\mathrm{col}} \mathcal{L}_{\mathrm{col}}$

    \State Backpropagate $\nabla_{\mathbf{q},\phi}\mathcal{L}$

    \State $\mathbf{q}$, $\phi$ $\gets$ Adam($\nabla_{\mathbf{q},\phi}$, $\eta_q$)

    \State $\mathbf{q}$ $\gets$ valid joint limits
    \State Geometric parameters in $\phi$ $\gets$ feasible ranges
\EndFor

\State \Return $\mathbf{q}, \phi$
\end{algorithmic}
\end{algorithm}

Given a target contact trajectory
\(
\mathbf{X}^{\star} = \{\mathbf{x}^{\star}_t\}_{t=1}^{T}
\),
we optimize the robot parameters by gradient descent through a differentiable forward kinematics model. Let
\(
\Theta = \{\mathbf{q}, \phi\}
\)
denote all optimizable parameters, where \(\mathbf{q}\) are the joint variables over the trajectory and \(\phi\) contains the design parameters, such as link lengths, pre-link lengths (the link added before attaching the joints), joint orientations, and optional mimic joint parameters. The forward model produces predicted fingertip positions
\[
\hat{\mathbf{X}} = g(\Theta),
\]
together with intermediate geometric quantities used for regularization and collision checking.

Our overall optimization problem can be written as
\begin{equation}
\Theta^{\star}
=
\arg\min_{\Theta}
\mathcal{L}_{\mathrm{track}}
+
\lambda_{\mathrm{joint}} \mathcal{L}_{\mathrm{joint}}
+
\lambda_{\mathrm{design}} \mathcal{L}_{\mathrm{design}}
+
\lambda_{\mathrm{col}} \mathcal{L}_{\mathrm{col}},
\end{equation}
where \(\mathcal{L}_{\mathrm{track}}\) is the trajectory tracking term, \(\mathcal{L}_{\mathrm{joint}}\) regularizes the joint trajectory, \(\mathcal{L}_{\mathrm{design}}\) regularizes the mechanism design, and \(\mathcal{L}_{\mathrm{col}}\) penalizes self-collision.

In our implementation, the tracking loss is an \(L_1\) distance between predicted and target fingertip trajectories,
\begin{equation}
\mathcal{L}_{\mathrm{track}}
=
\frac{1}{T}
\sum_{t=1}^{T}
\left\|
\hat{\mathbf{x}}_t - \mathbf{x}^{\star}_t
\right\|_{1},
\end{equation}
or equivalently the summed version when using sum reduction. We additionally impose temporal smoothness on the effective joint angles,
\begin{equation}
\mathcal{L}_{\mathrm{smooth}}
=
\frac{1}{T-1}
\sum_{t=1}^{T-1}
\left\|
\mathbf{q}^{\mathrm{eff}}_{t+1} - \mathbf{q}^{\mathrm{eff}}_{t}
\right\|_2^2,
\end{equation}
and include this term inside \(\mathcal{L}_{\mathrm{joint}}\). The remaining joint regularization terms softly keep the motion within feasible angular ranges. The design regularization term penalizes unnecessarily long links and, when mimic joints are enabled, also includes soft penalties on the mimic parameters for the Bennett linkage constraints. The total loss is then minimized with first-order gradient descent, followed by simple projection or clamping of parameters back to feasible ranges after each update.

\subsubsection{Collision module}
The collision between the links of the robot hand is crucial in the optimization framework. We formulate it as a differentiable framework where we compute collisions based on the link segment instead of the actual mesh of the robot hand. This speeds up the computation and creates a much better landscape compared with the raw mesh collision space. The algorithm is shown in Algorithm~\ref{alg:segment_collision}.

We use a simpler formulation in which each link is represented by its centerline segment. This produces a much smoother objective for first-order optimization while still discouraging self-intersections between the links. In our implementation, collision is therefore computed using pairwise distances between line segments extracted directly from the forward kinematics, rather than using full mesh geometry. 

Let segment \(i\) be defined by its endpoints
\[
\mathbf{p}_i^0,\ \mathbf{p}_i^1 \in \mathbb{R}^3,
\qquad
\mathbf{r}_i = \mathbf{p}_i^1 - \mathbf{p}_i^0,
\]
and segment \(j\) by
\[
\mathbf{q}_j^0,\ \mathbf{q}_j^1 \in \mathbb{R}^3,
\qquad
\mathbf{s}_j = \mathbf{q}_j^1 - \mathbf{q}_j^0.
\]
A point on each segment can be parameterized as
\begin{equation}
\mathbf{c}_i(u) = \mathbf{p}_i^0 + u \mathbf{r}_i,
\qquad
\mathbf{c}_j(v) = \mathbf{q}_j^0 + v \mathbf{s}_j,
\qquad
u,v \in [0,1].
\end{equation}
The unconstrained closest points on the supporting lines are obtained by minimizing
\begin{equation}
\left\|
\mathbf{c}_i(u) - \mathbf{c}_j(v)
\right\|_2^2.
\end{equation}
Defining
\begin{equation}
a = \mathbf{r}_i^\top \mathbf{r}_i,
\qquad
e = \mathbf{s}_j^\top \mathbf{s}_j,
\qquad
b = \mathbf{r}_i^\top \mathbf{s}_j,
\qquad
\mathbf{w}_0 = \mathbf{p}_i^0 - \mathbf{q}_j^0,
\qquad
c = \mathbf{r}_i^\top \mathbf{w}_0,
\qquad
d = \mathbf{s}_j^\top \mathbf{w}_0,
\end{equation}
the line-line minimizers are
\begin{equation}
u^{\star} = \frac{bd - ce}{ae - b^2},
\qquad
v^{\star} = \frac{ad - bc}{ae - b^2}.
\end{equation}
To obtain the closest points on the \emph{segments}, we clamp these parameters to the interval \([0,1]\),
\begin{equation}
\bar{u} = \mathrm{clip}(u^{\star}, 0, 1),
\qquad
\bar{v} = \mathrm{clip}(v^{\star}, 0, 1),
\end{equation}
and compute
\begin{equation}
\mathbf{c}_i^{\star} = \mathbf{p}_i^0 + \bar{u}\,\mathbf{r}_i,
\qquad
\mathbf{c}_j^{\star} = \mathbf{q}_j^0 + \bar{v}\,\mathbf{s}_j.
\end{equation}
The resulting segment-segment distance is
\begin{equation}
d_{ij}
=
\left\|
\mathbf{c}_i^{\star} - \mathbf{c}_j^{\star}
\right\|_2.
\end{equation}
The closest points are first solved on the infinite supporting lines, then clamped to the segment domain, and the Euclidean distance between the resulting points is used as the collision distance. 

We only evaluate this penalty on valid segment pairs. Self-pairs are removed, directly adjacent links are ignored, and segment pairs that share a joint endpoint are also excluded. In the tree structure, we additionally drop pairs that are adjacent across branch starts. Denoting the set of valid pairs by \(\mathcal{P}\), the collision loss is
\begin{equation}
\mathcal{L}_{\mathrm{col}}
=
\sum_{(i,j)\in\mathcal{P}}
\max\!\left(0,\ w - d_{ij}\right),
\end{equation}
where \(w\) is a clearance radius corresponding to the link dimension. Thus, pairs farther apart than \(w\) have no penalty, while pairs closer than \(w\) receive a linear penalty that pushes them apart. 

\begin{algorithm}[t]
\caption{\textsc{SegmentCollisionPenalty}: Differentiable collision penalty}
\label{alg:segment_collision}
\begin{algorithmic}[1]
\Require segment set $\mathcal{S}=\{(\mathbf{a}_i,\mathbf{b}_i)\}_{i=1}^{L}$ from forward kinematics, clearance radius $w$
\Ensure collision penalty $\mathcal{L}_{\mathrm{col}}$

\State Initialize valid-pair mask $\mathcal{M} \gets \mathbf{1}^{L \times L}$

\For{each pair of segments $(i,j)$}
    \If{$i=j$}
        \State $\mathcal{M}_{ij} \gets 0$ \Comment{ignore self-pairs}
    \ElsIf{$i$ and $j$ are adjacent links}
        \State $\mathcal{M}_{ij} \gets 0$ \Comment{ignore neighboring links}
    \ElsIf{$i$ and $j$ share a joint endpoint}
        \State $\mathcal{M}_{ij} \gets 0$ \Comment{ignore topologically connected pairs}
    \EndIf
\EndFor

\State $\mathcal{L}_{\mathrm{col}} \gets 0$
\For{each valid pair $(i,j)$ with $\mathcal{M}_{ij}=1$}
    \State $d_{ij} \gets \textsc{ClosestSegmentDistance}\big((\mathbf{a}_i,\mathbf{b}_i),(\mathbf{a}_j,\mathbf{b}_j)\big)$
    \State $\ell_{ij} \gets \max(0,\, w - d_{ij})$
    \State $\mathcal{L}_{\mathrm{col}} \gets \mathcal{L}_{\mathrm{col}} + \ell_{ij}$
\EndFor

\State \Return $\mathcal{L}_{\mathrm{col}}$
\end{algorithmic}
\end{algorithm}

\subsection{Training pipeline}\label{sec:appendix_train_actor}
As shown in Algorithm~\ref{alg:best_of_k_actor}, we train a trajectory-conditioned stochastic actor to learn to predict the design and joint angle initialization parameters. For a sampled contact trajectory \(\mathbf{X}\), we first compute a context vector \(\mathbf{c}\), from a pretrained trajectory encoder based on the dataset. The actor \(f_{\theta}\) maps this context to a deterministic anchor action
\[
\boldsymbol{\mu}_{\theta}(\mathbf{c}) = f_{\theta}(\mathbf{c}) \in \mathbb{R}^{A},
\]
where \(A\) is the action dimension. Rather than predicting a full covariance, we use a fixed exploration variance of \(\sigma>0\) and sample \(K\) candidate actions as
\[
\mathbf{a}_k = \boldsymbol{\mu}_{\theta}(\mathbf{c}) + \sigma \boldsymbol{\epsilon}_k,
\qquad
\boldsymbol{\epsilon}_k \sim \mathcal{N}(\mathbf{0}, \mathbf{I}),
\qquad
k=1,\dots,K.
\]
Each sampled action is decoded into design parameters and initial joint angles, then evaluated by running the co-design optimization in Algorithm~\ref{alg:joint_design_opt} for a small number of epochs. This produces a final fitting loss \(\ell_k\), a Bennett collision penalty \(b_k\), and an angle-consistency penalty \(a_k\). We convert these quantities into a scalar reward
\[
r_k = s^{\mathrm{col}}_k\, s^{\mathrm{angle}}_k\, s^{\mathrm{loss}}_k,
\]
where
\[
s^{\mathrm{bennett}}_k = \sigma_g\!\left(\frac{8.0 - b_k}{2.0}\right),\qquad
s^{\mathrm{angle}}_k = \sigma_g\!\left(\frac{0.1 - 100\,a_k}{0.025}\right),\qquad
s^{\mathrm{loss}}_k = \sigma_g\!\left(\frac{100.0 - \ell_k}{25.0}\right),
\]
and \(\sigma_g(\cdot)\) denotes the logistic sigmoid. We then select the single best sampled action
\[
k^\star = \arg\max_{k} r_k,
\qquad
\mathbf{a}^{\star} = \mathbf{a}_{k^\star},
\]
and train the actor by regressing its predicted mean toward this best action:
\[
\mathcal{L}_{\mathrm{actor}}
=
\left\|
\boldsymbol{\mu}_{\theta}(\mathbf{c}) - \mathbf{a}^{\star}
\right\|_2^2.
\]
Thus, the actor repeatedly samples a small batch of noisy candidates around its current prediction and learns to move its mean toward the highest-reward sample.

\begin{algorithm}[t]
\caption{Training the trajectory-conditioned actor}
\label{alg:best_of_k_actor}
\begin{algorithmic}[1]
\Require trajectory dataset \(\mathcal{D}\), actor network \(f_\theta\), fixed noise scale \(\sigma\), number of samples \(K\), number of episodes \(E\)
\For{episode \(e = 1,\dots,E\)}
    \State Sample one trajectory \(\mathbf{X} \sim \mathcal{D}\)
    \State Apply smooth trajectory augmentation to \(\mathbf{X}\) with random seed \(s_e\)
    \State Encode trajectory context \(\mathbf{c} \gets \mathrm{Encode}(\mathbf{X})\)
    \State Compute mean action \(\boldsymbol{\mu} \gets f_\theta(\mathbf{c})\)

    \For{\(k = 1,\dots,K\)}
        \State Sample noise \(\boldsymbol{\epsilon}_k \sim \mathcal{N}(\mathbf{0}, \mathbf{I})\)
        \State Form candidate action
        \[
        \mathbf{a}_k \gets \boldsymbol{\mu} + \sigma \boldsymbol{\epsilon}_k
        \]
        \State Decode \(\mathbf{a}_k\) into design parameters and initial joint angles
        \State Run inner optimization / evaluator on \(\mathbf{a}_k\)
        \State Obtain final loss \(\ell_k\), Bennett penalty \(b_k\), angle penalty \(a_k\)
        \State Compute reward
        \[
        r_k \gets
        \sigma_g\!\left(\frac{8.0 - b_k}{2.0}\right)
        \cdot
        \sigma_g\!\left(\frac{0.1 - 100 a_k}{0.025}\right)
        \cdot
        \sigma_g\!\left(\frac{100.0 - \ell_k}{25.0}\right)
        \]
    \EndFor

    \State Select best candidate
    \[
    k^\star \gets \arg\max_k r_k,
    \qquad
    \mathbf{a}^\star \gets \mathbf{a}_{k^\star}
    \]
    \State Recompute mean action \(\boldsymbol{\mu} \gets f_\theta(\mathbf{c})\)
    \State Compute actor loss
    \[
    \mathcal{L}_{\mathrm{actor}} \gets \|\boldsymbol{\mu} - \mathbf{a}^\star\|_2^2
    \]
    \State Update \(\theta\) with Adam on \(\mathcal{L}_{\mathrm{actor}}\)
    \State Clip gradients and log best / mean reward statistics
\EndFor
\end{algorithmic}
\end{algorithm}

\subsection{Mechanism generation framework}\label{sec:appendix_mechanism_gen}

Given an optimized design and a selected joint-angle combination, we generate the mesh of this mechanism by recursively instantiating rigid links and Bennett four-bar linkages. Let
\[
\phi =
\left\{
\ell_j,\;
{\ell}_b,\;
R_j,\;
m_j,\;
\delta^{\mathrm{off}}_j
\right\}
\]
denote the optimized design parameters, where \(\ell_j\) is a link length, \({\ell}_b\) is a pre-link length for branch \(b\), \(R_j \in SO(3)\) is the fixed orientation of joint \(j\), \(m_j\) is the Bennett ratio for the \(j\)-th mimic joint, and \(\delta^{\mathrm{off}}_j\) is the non-major phase offset used to initialize the Bennett construction. We denote the selected joint-angle vector by
\[
\mathbf{q} = [q_0,\dots,q_{J-1}]^\top .
\]

We use homogeneous transforms
\[
\mathbf{T}(\mathbf{R},\mathbf{t}) =
\begin{bmatrix}
\mathbf{R} & \mathbf{t} \\
\mathbf{0}^\top & 1
\end{bmatrix},
\qquad
\mathbf{R}(\theta) \in SO(3)
\]
to place each component in the world frame. The mechanism is constructed from a root chain followed by one or more branches.

\paragraph{Root chain.}
The root frame is initialized at the wrist,
\[
\mathbf{T}^{\mathrm{root}}_{0} = \mathbf{I}_4.
\]
A fixed pre-root orientation is first applied,
\[
\mathbf{T}_{\mathrm{pre}} =
\mathbf{T}(R_0,\mathbf{0}),
\]
followed by a translation along the local negative \(x\)-axis by the pre-link length \({\ell}_0\), and then the first revolute joint rotation:
\begin{equation}
\mathbf{T}_0
=
\mathbf{T}^{\mathrm{root}}_{0}
\mathbf{T}(R_0,\mathbf{0})
\mathbf{T}(\mathbf{I},-{\ell}_0 \mathbf{e}_x)
\mathbf{T}(\mathbf{R}_z(q_0),\mathbf{0}).
\end{equation}
The first physical bar is then rendered as a straight link of length \(\ell_0\) aligned with the local negative \(x\)-axis.

\paragraph{Branch pre-link.}
Each branch \(b\) starts from the same base frame after the root link and attaches a fixed pre-link orientation followed by a branch-specific translation and a revolute joint:
\begin{equation}
\mathbf{T}^b_0
=
\mathbf{T}^{\mathrm{base}}
\mathbf{T}(R_{j_b},-\ell_0 \mathbf{e}_x)
\mathbf{T}(\mathbf{I},-{\ell}_b \mathbf{e}_x)
\mathbf{T}(\mathbf{R}_z(q_{j_b}),\mathbf{0}),
\end{equation}
where \(j_b\) is the first joint index of branch \(b\).

\paragraph{Bennett linkage generation.}
For each mimic joint, we generate a local Bennett 4R linkage from the current design parameters. Let
\[
b_j = \ell_j,
\qquad
a_j = |m_j|\, b_j,
\qquad
q_j^{\mathrm{init}} = q_j + \delta^{\mathrm{off}}_j.
\]
The first two hinge axes are fixed from the current local frame and the stored joint orientation:
\[
\mathbf{p}_0 = \mathbf{0},
\qquad
\mathbf{u}_0 = \mathbf{e}_z,
\]
\[
\mathbf{p}_1 = R_{j+1}(-b_j \mathbf{e}_x),
\qquad
\mathbf{u}_1 = R_{j+1}\mathbf{e}_z.
\]
The fourth hinge anchor is initialized as
\[
\mathbf{p}_3 = \mathbf{R}_z(-q_j^{\mathrm{init}})(-a_j \mathbf{e}_x).
\]
The remaining Bennett variables are
\[
\mathbf{x}
=
\begin{bmatrix}
\mathbf{p}_2 \\
\mathbf{u}_2 \\
\mathbf{u}_3
\end{bmatrix}
\in \mathbb{R}^{9},
\]
which are obtained by nonlinear least squares.

Define the link bar vectors
\[
\mathbf{e}_{21} = \mathbf{p}_2 - \mathbf{p}_1,
\qquad
\mathbf{e}_{23} = \mathbf{p}_2 - \mathbf{p}_3,
\qquad
\mathbf{e}_{30} = \mathbf{p}_3 - \mathbf{p}_0,
\]
and the twist sine/cosine terms
\[
s_{ij} = \|\mathbf{u}_i \times \mathbf{u}_j\|_2,
\qquad
c_{ij} = \mathbf{u}_i^\top \mathbf{u}_j.
\]
The Bennett residual is formed from orthogonality, link length, twist between opposite links, and ratio constraints:
\begin{equation}
\mathbf{r}_{\mathrm{orth}} =
\frac{1}{\max(a_j,b_j,\varepsilon)}
\begin{bmatrix}
\mathbf{e}_{21}^\top \mathbf{u}_1 \\
\mathbf{e}_{21}^\top \mathbf{u}_2 \\
\mathbf{e}_{23}^\top \mathbf{u}_2 \\
\mathbf{e}_{23}^\top \mathbf{u}_3 \\
\mathbf{e}_{30}^\top \mathbf{u}_3 \\
\mathbf{e}_{30}^\top \mathbf{u}_0
\end{bmatrix},
\end{equation}
\begin{equation}
\mathbf{r}_{\mathrm{len}} =
\begin{bmatrix}
(\|\mathbf{e}_{21}\|_2-a_j)/a_j \\
(\|\mathbf{e}_{23}\|_2-b_j)/b_j \\
(\|\mathbf{e}_{30}\|_2-a_j)/a_j
\end{bmatrix},
\end{equation}
\begin{equation}
\mathbf{r}_{\mathrm{twist}} =
\begin{bmatrix}
s_{01}-s_{23} \\
c_{01}-c_{23} \\
s_{12}-s_{30} \\
c_{12}-c_{30}
\end{bmatrix},
\qquad
r_{\mathrm{ratio}} = b_j s_{12} - a_j s_{01}.
\end{equation}
We also softly enforce unit-norm axes with
\[
\mathbf{r}_{\mathrm{unit}} =
\begin{bmatrix}
\|\mathbf{u}_2\|_2 - 1 \\
\|\mathbf{u}_3\|_2 - 1
\end{bmatrix}.
\]
The full Bennett bar generation solution is
\begin{equation}
\mathbf{x}^{\star}
=
\arg\min_{\mathbf{x}}
\left\|
w_{\mathrm{orth}} \mathbf{r}_{\mathrm{orth}}
\oplus
w_{\mathrm{len}} \mathbf{r}_{\mathrm{len}}
\oplus
w_{\mathrm{twist}} \mathbf{r}_{\mathrm{twist}}
\oplus
w_{\mathrm{ratio}} r_{\mathrm{ratio}}
\oplus
\mathbf{r}_{\mathrm{unit}}
\right\|_2^2,
\end{equation}
where \(\oplus\) denotes concatenation. We initialize this optimization from multiple ring samples around the known axis \(\mathbf{u}_1\), keep the lowest-cost candidates, and select the final solution whose \(\mathbf{p}_2\) lies farthest from \(\mathbf{p}_0\), which disambiguates the mirrored Bennett branches.

\paragraph{Frame propagation and rendering.}
Once \((\mathbf{p}_0,\mathbf{u}_0),\dots,(\mathbf{p}_3,\mathbf{u}_3)\) are obtained, we construct four joint frames whose local \(z\)-axes coincide with the hinge axes and whose \(x\)-axes are chosen by continuity. For the mimic joint $j$, let these four axes of the four-bar local transforms be
\[
{\mathbf{\hat T}}_{j,0},{\mathbf{\hat T}}_{j,1},{\mathbf{\hat T}}_{j,2},{\mathbf{\hat T}}_{j,3}.
\]
The next mechanism frame is taken from the second Bennett joint frame and composed with the child revolute angle:
\begin{equation}
\hat{\mathbf{T}}_{j+1}
=
\hat{\mathbf{T}}_{j,1}
\mathbf{T}(\mathbf{R}_z(q_{j+1}),\mathbf{0}),
\qquad
\mathbf{T}_{j+1}
=
\mathbf{T}_{j}\hat{\mathbf{T}}_{j+1}.
\end{equation}
The four Bennett links are then rendered as rigid bars connecting consecutive joint positions. At the end of a branch, if no further mimic block exists, we append a terminal bar of length \(\ell_j\) along the current local negative \(x\)-axis. Its tip is
\begin{equation}
\mathbf{p}_{\mathrm{tip}}
=
\mathbf{T}_{j}
\begin{bmatrix}
-\ell_j \\ 0 \\ 0 \\ 1
\end{bmatrix}.
\end{equation}

Overall, we convert the optimized design variables into an explicit articulated mechanism consisting of a root link, branch pre-links, Bennett 4R linkages, and terminal bars, all represented as rigid meshes and world-frame transforms.

\subsection{Parameter list}
\begin{table}[t]
\centering
\caption{\textbf{Hyperparameters for differentiable co-design optimization.}}
\label{tab:opt_hparams}
\small
\begin{tabular}{llc}
\toprule
\textbf{Category} & \textbf{Hyperparameter} & \textbf{Value} \\
\midrule
Optimization & Joint-angle optimizer & AdamW \\
Optimization & Design optimizer (link / pre-link) & AdamW \\
Optimization & Design optimizer (joint orientation) & AdamW \\
Optimization & Mimic-parameter optimizer & AdamW \\
Optimization & Joint learning rate $lr$ & $5\times 10^{-3}$ \\
Optimization & Design learning rate $lr_{\mathrm{design}}$ & $5\times 10^{-3}$ \\
Optimization & Orientation optimizer learning rate & $0.5\,lr_{\mathrm{design}}$ \\
Optimization & Mimic optimizer learning rate & $0.5\,lr_{\mathrm{design}}$ \\
Optimization & Weight decay & $0$ \\
Scheduler & LR scheduler & StepLR \\
Scheduler & Step size & $1000$ \\
Scheduler & Decay factor $\gamma$ & $0.9$ \\
Iterations & Default optimization steps (CLI) & $2\times 10^{4}$ \\
Iterations & Inner optimization steps in actor evaluation & $500$ \\
\midrule
Loss weights & Tracking loss weight & $10^{4}$ \\
Loss weights & Collision loss weight & $10^{4}$ \\
Loss weights & Joint regularization weight & $10^{-2}$ \\
Loss weights & Design length regularization weight & $10^{-6}$ \\
Loss weights & Mimic loss weight & $1$ \\
Loss weights & Smoothness coefficient & $10^{-2}$ \\
\midrule
Parameter bounds & Minimum link length & $0.025$ m \\
Parameter bounds & Maximum link length & $0.15$ m \\
Parameter bounds & Minimum mimic link length & $0.02$ m \\
Parameter bounds & Minimum joint-angle margin & $0.3$ rad \\
Parameter bounds & Mimic residual lower bound & $10^{-3}$ \\
\bottomrule
\end{tabular}
\end{table}

\begin{table}[t]
\centering
\caption{\textbf{Hyperparameters for trajectory-conditioned actor training.} The actor predicts a mean action and samples candidates using fixed Gaussian noise.}
\label{tab:actor_hparams}
\small
\begin{tabular}{llc}
\toprule
\textbf{Category} & \textbf{Hyperparameter} & \textbf{Value} \\
\midrule
Architecture & Network type & MLP \\
Architecture & Hidden width & $256$ \\
Architecture & Number of hidden layers & $3$ \\
Architecture & Activation & SiLU \\
Architecture & Weight initialization & Xavier uniform (gain $0.5$) \\
Architecture & Bias initialization & $0$ \\
\midrule
Training & Optimizer & Adam \\
Training & Actor learning rate & $2\times 10^{-4}$ \\
Training & Gradient clipping norm & $5.0$ \\
Training & Number of episodes & $5000$ \\
Training & Samples per episode $K$ & $8$ (default) \\
Training & Fixed sampling standard deviation & $1.0$ \\
Training & Random seed & $42$ \\
\midrule
Parallel evaluation & Workers per GPU & $2$ \\
\midrule
Input context & Statistics feature & first, mean, std of contact trajectory \\
Input context & Optional latent encoder & pretrained trajectory autoencoder \\
\bottomrule
\end{tabular}
\end{table}

\begin{table}[t]
\centering
\caption{\textbf{Trajectory augmentation and reward hyperparameters used for actor training.}}
\label{tab:reward_hparams}
\small
\begin{tabular}{llc}
\toprule
\textbf{Category} & \textbf{Hyperparameter} & \textbf{Value} \\
\midrule
Augmentation & Augmentation probability & $0.7$ \\
Augmentation & Smooth offset standard deviation & $0.0015$ m \\
Augmentation & Number of augmentation knots & $6$ \\
Augmentation & Offset type & shared smooth 3D trajectory offset \\
\midrule
Reward & Bennett penalty sigmoid center & $8.0$ \\
Reward & Bennett penalty sigmoid scale & $2.0$ \\
Reward & Angle penalty sigmoid center & $0.1$ \\
Reward & Angle penalty sigmoid scale & $0.025$ \\
Reward & Angle penalty scaling before sigmoid & $\times 10^{2}$ \\
Reward & Final loss sigmoid center & $100.0$ \\
Reward & Final loss sigmoid scale & $25.0$ \\
Reward & Reward aggregation & product of three sigmoid terms \\
\midrule
Evaluation frames & Frames for Bennett / angle evaluation & frame $0$ + $2$ random frames \\
\bottomrule
\end{tabular}
\end{table}

\end{document}